\documentclass{IEEEtran}

\usepackage{amsthm,amsmath}
\usepackage[utf8]{inputenc} 
\usepackage{amsmath}
\usepackage{amssymb}
\usepackage{placeins}
\usepackage{url}
\usepackage{xcolor}
\usepackage{booktabs}
\usepackage{makecell}
\usepackage{bm}
\usepackage{scalerel}
\usepackage{array}
\usepackage{calc}
\usepackage{graphicx}
\usepackage{tasks}
\usepackage{comment}
\usepackage{microtype}
\usepackage{multirow}
\usepackage{hyperref}
\usepackage{tikz}

\usetikzlibrary{svg.path}

\definecolor{orcidlogocol}{HTML}{A6CE39}
\tikzset{
	orcidlogo/.pic={
		\fill[orcidlogocol] svg{M256,128c0,70.7-57.3,128-128,128C57.3,256,0,198.7,0,128C0,57.3,57.3,0,128,0C198.7,0,256,57.3,256,128z};
		\fill[white] svg{M86.3,186.2H70.9V79.1h15.4v48.4V186.2z}
		svg{M108.9,79.1h41.6c39.6,0,57,28.3,57,53.6c0,27.5-21.5,53.6-56.8,53.6h-41.8V79.1z M124.3,172.4h24.5c34.9,0,42.9-26.5,42.9-39.7c0-21.5-13.7-39.7-43.7-39.7h-23.7V172.4z}
		svg{M88.7,56.8c0,5.5-4.5,10.1-10.1,10.1c-5.6,0-10.1-4.6-10.1-10.1c0-5.6,4.5-10.1,10.1-10.1C84.2,46.7,88.7,51.3,88.7,56.8z};
	}
}

\newcommand\orcidicon[1]{\href{https://orcid.org/#1}{\mbox{\scalerel*{
				\begin{tikzpicture}[yscale=-1,transform shape]
				\pic{orcidlogo};
				\end{tikzpicture}
			}{|}}}}

\begin{document}
\newcommand{\autocite}[1]{\cite{#1}}
\newcommand{\csentence}[1]{\textbf{#1}}
\newcommand{\parttitle}[1]{\par\noindent\textbf{#1}}

\title{Scientific Image Tampering Detection Based On Noise Inconsistencies: A Method And Datasets}

\author{Ziyue Xiang \orcidicon{0000-0001-6054-5801}\,, Daniel E. Acuna \orcidicon{0000-0002-7765-1595}}

\maketitle

\begin{abstract}
Scientific image tampering is a problem that affects not only authors but also the general perception of the research community. Although previous researchers have developed methods to identify tampering in natural images, these methods may not thrive under the scientific setting as scientific images have different statistics, format, quality, and intentions. Therefore, we propose a scientific-image specific tampering detection method based on noise inconsistencies, which is capable of learning and generalizing to different fields of science. We train and test our method on a new dataset of manipulated western blot and microscopy imagery, which aims at emulating problematic images in science. The test results show that our method can detect various types of image manipulation in different scenarios robustly, and it outperforms existing general-purpose image tampering detection schemes. We discuss applications beyond these two types of images and suggest next steps for making detection of problematic images a systematic step in peer review and science in general.
\end{abstract}

\begin{IEEEkeywords}
Scientific images, Digital image forensics, Noise inconsistency, Scientific image manipulation dataset
\end{IEEEkeywords}

\newcolumntype{C}[1]{>{\centering\let\newline\\\arraybackslash\hspace{0pt}}m{#1}}

\section{Introduction}


The use of digital images has become increasingly ubiquitous in all types of publications. What comes with the growing importance of digital images is the development of image tampering techniques. In the past, modifying or concealing the content of an image would require dedicated personnel and tools. Today, however, image tampering is much easier with state-of-the-art image processing software. This trend has affected many aspects of our society, as we see prominent forgery cases occur in journalism and academia \cite{farid2016photo}. Consequently, many detection techniques have been developed for these scenarios (see \cite{ISI:000265093400004}). Only recently, however, attention has been paid to image manipulation in scientific publications \cite{gilbert2009science}. Although it is possible to use existing methods on scientific images directly, we hypothesize that significant adaptations must be made due to the fact that they usually possess distinctive statistical patterns, formats and resolutions. In this work, we aim at developing a scientific-specific image manipulation detection technique, which we test on a novel scientific image manipulation dataset of western blots and microscopy imagery---there are no datasets openly available about scientific image manipulation yet (but see \cite{beck2016shaping}). Thus, as most scientific images increasingly come in digital form, the detection of possible manipulations should also get at the same level of quality as other fields that use digital images only.

It is undeniable that an increasing amount of tampered images are finding their ways into scientific publications. Bik, Casadevall and Fang \cite{bik2016prevalence} examined 20,621 biomedical research papers from 1995 to 2014, where they find that at least 1.9 percent are subject to deliberate image manipulation. The fact that these suspicious papers went through the careful reviewing process suggests how difficult it is to examine image tampering in scientific research manually. Because the large quantity of digital images present in submitted manuscript, it is be crucial for publishers to be able to identify image manipulation in an automated fashion.

The scientific research context sets a different tolerance for image manipulation. Many operations, including resizing, contrast adjusting, sharpening, and white balancing are generally acceptable as part of the figure preparation process. However, some others types of tampering, especially the ones that alter the image content semantically, are strictly prohibited. These manipulations include copy-move (without proper attribution), splicing, removal, and retouching\footnote{\url{https://ori.hhs.gov/education/products/RIandImages/guidelines/list.html}}. Acuna, Brookes and Kording \cite{acuna2018bioscience} developed a method to detect figure element reuse across a paper database. Intra-image copy-move can be detected rather robustly with SIFT features and pattern matching \autocite{huang2008detection}. However, detection of image manipulation that does not involve reuse is significantly more challenging. A comprehensive scientific image manipulation detection pipeline should include manipulation detection.

As scientific papers are reviewed by experts, we reckon that articles containing manipulations that incur in contextual inconsistencies (e.g., brain activation patterns from fMRI in the middle of a microscopy image) will be easily picked out. What humans \emph{cannot} see properly is the noise pattern within an image---and scientists seeking to falsify images exploit this weakness. Therefore, we propose a novel image tampering detection method for scientific images, which is based on uncovering noise inconsistencies. Specifically, our proposed method contains the following features:
\begin{enumerate}
	\item It is based on supervised learning, which is capable of learning from existing databases and new instances.
	\item It works for images of different resolutions and from different devices.
	\item It is not restricted to any specific image format.
	\item It is capable of generating good predictions with a small training set.
	\item It is flexible and can be fine-tuned for different fields of science.
\end{enumerate}

In section \ref{sec::related_works}, we briefly summarize previous work on digital image forensics. In section \ref{sec::overall_design}, we discuss the design of our proposed method. In section \ref{sec::experiments}, we introduce our scientific image manipulation datasets and present the test results of our method on them. In section \ref{sec::conclusion}, we conclude by discussing limitations and future extension of our method.

\section{Previous Work}\label{sec::related_works}

There have been a large amount of previous research on image tampering detection, but very few of them focus on scientific images. The first class of tampering detection methods aims at detecting a specific type of manipulation, the most common being resizing and resampling \autocite{popescu2005exposing, kirchner2008fast, dalgaard2010role, feng2012normalized, mahdian2008blind}, median filtering \autocite{kirchner2010detection, kang2013robust, cao2010forensic, chen2011median}, contrast enhancement \autocite{stamm2010forensic1, yao2009detect, stamm2008blind, stamm2010forensic2}, blurring \autocite{liu2008image}, and multiple JPEG compression \autocite{bianchi2011detection, bianchi2012image, neelamani2006jpeg, qu2008convolutive}. Many of these manipulations are valid in the scientific research context, and it can be non-trivial to merge results from single detectors in order to build a comprehensive one.

The second class of tampering detection methods aims at general-purpose image tampering detection. Dirik and Memon \cite{dirik2009image} try to catch the inconsistency of Color Filtering Array (CFA) patterns within images taken by digital cameras---a signal generated by digital cameras. However, scientific images are not necessarily taken by digital cameras. Wang, Dong and Tan \cite{wang2014exploring} leverage the characteristics of the DCT coefficients in JPEG images to achieve tampering localization, but the method is confined to a specific format. Mahdian and Saic \cite{mahdian2009using} propose a method that predicts tampered regions based on wavelet transform and noise level estimation. All these methods are unable to learn from data, which limits their abilities to generalize to different fields. Another group of methods combines steganalysis tools \autocite{fridrich2012rich, pevny2010steganalysis} with Gaussian Mixture Models (GMM) to identify potentially manipulated regions  \autocite{fan2015general, cozzolino2015splicebuster}. These unsupervised-learning-based methods are also unable to learn from existing database effectively and therefore tend to underperform in practice. 

Because of the occurrence of large image datasets, neural-network-based tampering detection methods are likely to yield good performance \autocite{bappy2017exploiting}, especially those based on Convolutional Neural Networks (CNN) \autocite{bayar2016deep, bayar2018constrained, zhou2018learning}. They usually target high resolution natural images. It is unclear, however, whether they can be transitioned for the scientific scenario. For example, it is challenging to train such a network for scientific images exclusively as they usually require tens of thousands of images as training data, which to the best of our knowledge is not yet available.

\section{Our Proposed Method}\label{sec::overall_design}

\begin{figure*}[htpb]
	\centering
	\includegraphics[width=0.9\linewidth]{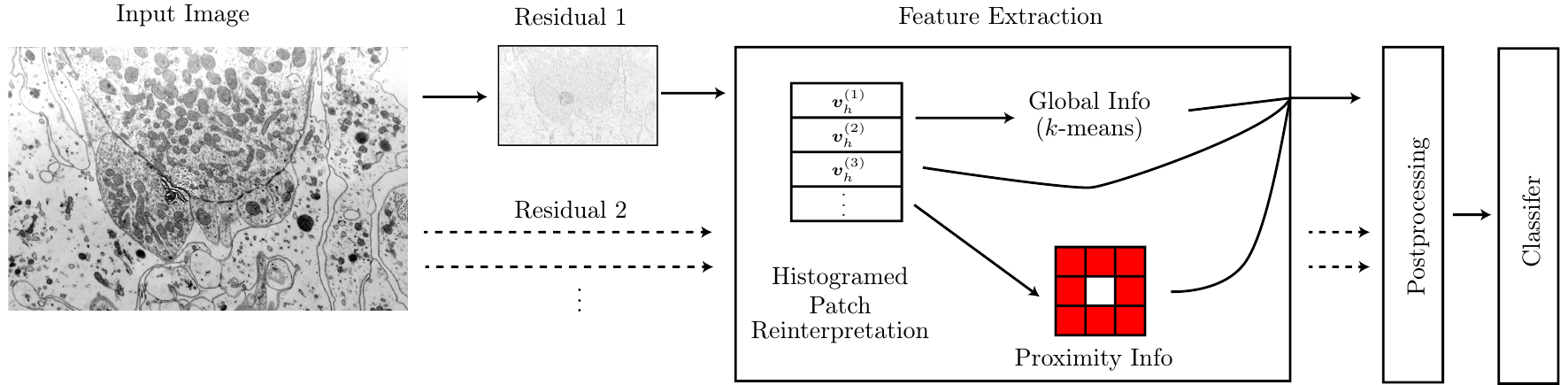}
	\caption{\csentence{Overall design of our proposed method.} The input image goes through several residual generators and feature extractors in parallel. All extracted features will be merged in a postprocessing step and then fed to a classifier.}
	\label{fig::method_work_flow}
\end{figure*}

Our method is based on a combination of several heterogeneous feature extractors that are later combined to produce single predictions for patches (Figure \ref{fig::method_work_flow}). At first, an input image will go through a variable amount of residual image generators. The type and amount of these generators can be chosen based on the application. Each type of residual image will have its own feature extractor, which is based on our proposed feature extraction scheme with (possibly) different configurations. The features are then fed into a classifier after post-processing.

The proposed method works on residual images, which are essentially image after filtering or the difference between an image and its interpolated version. It is a way to discard content and emphasize noise pattern within an image, which is widely used in image manipulation detection practice. However, in many previous works, only one type of residuals is used \cite{dirik2009image, cozzolino2015splicebuster, zhou2018learning}. Because each residual may have different sensitivity levels to different types of manipulation, using only one not only limits the method's ability to detect a wide variety of manipulation, but also renders the method more vulnerable against adversaries. Therefore, we decide to combine a number of residuals in our method to increase the robustness.

Because our feature extraction method drastically reduces the dimensionality of image data, which relieves the need of a huge amount of training data, it is possible to use a light-weight classifier as the back end, such as logistic regression or support vector machine (SVM). As there are many ways to generate residual images, and that the feature extraction method comes with a number of parameters to decide, our image manipulation detection method possesses high degree of flexibility. Unlike the parameters in neural networks, for example, which are rather obscure for human beings, the underlying meanings of the parameters in our feature extraction method are straightforward. Therefore, it is easier for one to manually adapt our method for different fields. 



\subsection{Residual Image Generators}\label{sec::choice_of_filters}

There are numerous ways of generating residual images, we list the following ones because they are functional for a wide range of applications. Note that the capability of our method is significantly influenced by the choice of residuals. However, it is possible to design new residual image generators for specific scenarios.

\begin{figure}[htpb]
	\centering
	\includegraphics[width=0.95\linewidth]{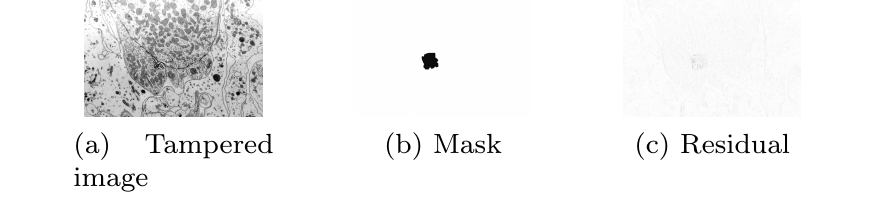}
	\caption{\csentence{Demonstration of steganalytic residual.}}
	\label{fig::stega_res_demo}
\end{figure}

\begin{figure}[htpb]
	\centering
	\includegraphics[width=0.95\linewidth]{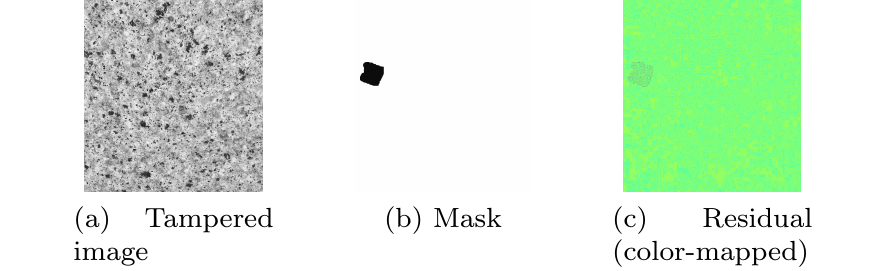}
	\caption{\csentence{Demonstration of ELA residual.}}
	\label{fig::ela_res_demo}
\end{figure}

\begin{figure}[htpb]
	\centering
	\includegraphics[width=0.95\linewidth]{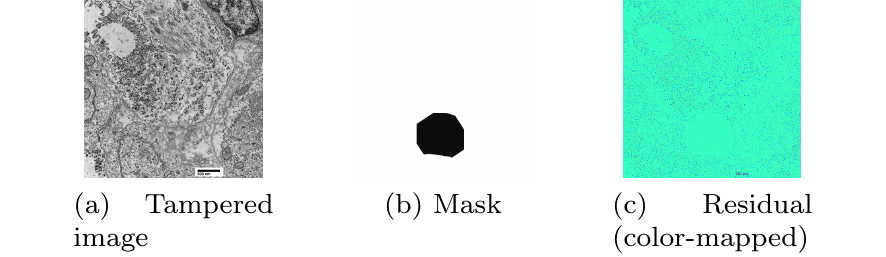}
	\caption{\csentence{Demonstration of median filtering residual.}}
	\label{fig::mf_res_demo}
\end{figure}

\begin{figure}[htpb]
	\centering
	\includegraphics[width=0.95\linewidth]{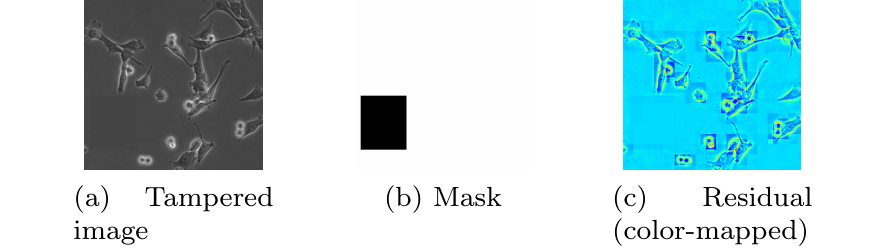}
	\caption{\csentence{Demonstration of wavelet denoising residual.}}
	\label{fig::wavelet_res_demo}
\end{figure}

\begin{enumerate}
	\item Steganalytic Filters
	
	Steganalysis (techniques used for detecting hidden messages in communications) has been used in image tampering detection practice extensively. This type of analysis aims to expose hidden information planted in images by steganography techniques. Although it is not directly linked to image tampering detection, it is suggested that that the tasks of image forensics and steganalysis are very much alike when the action of data embedding in steganography is treated as image manipulating \cite{qiu2014universal}. 
	Similar to the rich model strategy proposed in \autocite{fridrich2012rich}, we can apply many different filters and see which one can spot inconsistencies. In our work, we use several filters that provide a relatively comprehensive view of potential inconsistencies (Figure \ref{fig::high_pass_filters}).

	\begin{figure}[h!]
		\centering
		\includegraphics[width=0.8\linewidth]{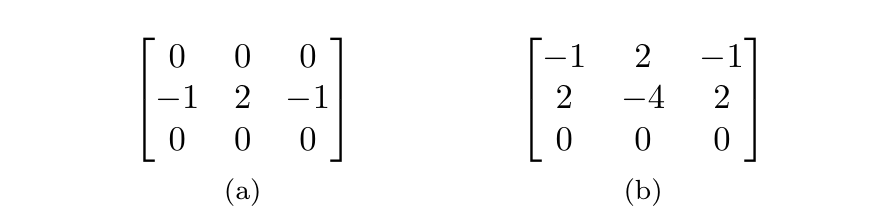}
		\caption{\csentence{High-pass filters selected in our experiment.}}
		\label{fig::high_pass_filters}
	\end{figure}
	
	The filters selected are high-pass because we want to throw away information about the image content and emphasize noise patterns as much as possible. The residual image in this case is the image after convolution. An example of steganalytic filtering residual is shown in Figure \ref{fig::stega_res_demo}.

	\item Error Level Analysis (ELA)

	ELA is an analysis technique that targets JPEG compression. The idea behind it is that the amount of error introduced by JPEG compression is nonlinear: a 90-quality JPEG image resaved at quality 90 is equivalent to a one-time save of quality 81; a 90-quality JPEG image resaved at quality 75 is equivalent to a one-time save of quality 67.5 \autocite{krawetz2007picture}; and so on. If some part of a JPEG-compressed image is altered with a different JPEG quality factor, when it is compressed again, the loss of information of that part will differ from other regions. To uncover the inconsistency, ELA residual is computed by intentionally resaving the image in JPEG format with a particular quality (e.g. 90) and then computing the difference of the two images. An example of ELA residual is shown in Figure \ref{fig::ela_res_demo}.
	
	\item Median Filtering Residual
	
	Median filtering can suppress the noise of an image. When applying median filtering to a tampered image, the tampered part may possess a different noise pattern and therefore respond differently. The median filtering residual is the difference between the original image and median filtered image. An example is shown in Figure \ref{fig::mf_res_demo}.
	
	\item Wavelet Denoising Residual
	
	Wavelet denoising is a type of denoising method that represents an image in wavelet domain and cancels the noise based on that representation. Similar to the median filtering residual's case, the tampered region may react differently compared to the rest of the image and therefore give away its own identity. It is also suggested by Dirik and Memon \cite{dirik2009image} that using wavelet denoising can uncover the sensor noise inconsistency of digital cameras. The wavelet denoising residual is given by the difference between the original image and the denoised image. An example is shown in Figure \ref{fig::wavelet_res_demo}.

\end{enumerate}

It is worth noticing that the tampered images in the demonstrations are selected so that the manipulation pattern is visible in the specific residual. However, in practice, this may not always be the case. Usually it is necessary to examine multiple residual images before drawing a conclusion. 

\subsection{Feature Extraction}\label{sec::feature_extraction}

Our method is patch-based, which means it will generate a prediction for each patch in the image. Using patches instead of single pixels to represent an image not only shrinks the scale of computation, but also enriches the amount of statistical information within each smallest unit. At the limit, the patch size can be chosen so that pixel-based and patch-based become almost the same. After deciding on the patch size, the feature extraction step will generate a corresponding feature vector for each patch in the image. In this section, we discuss how these features vectors are computed.

\subsubsection{Patch Reinterpretation}

Residuals reduce the complexity of image data, but they still have the same dimensionality as the original image. To further compress data for classification, we propose a new feature extraction method for image tampering detection. Intuitively, an image region is considered to be \emph{tampered} not because it is unique itself, but mainly due to the fact that it is different from \emph{the rest of the image}. Therefore, an ideal feature design should contain sufficient amount of global information. We add global information by reinterpreting an image region using the rest of the image.

\begin{table}[htpb]
	\caption{List of symbols used in feature extraction}
	\centering{\scriptsize
		\begin{tabular}{cm{0.45\linewidth}}
			\toprule
			Symbol & \multicolumn{1}{c}{Description} \tabularnewline \midrule
			$(h, w)$ & size of the image \tabularnewline \midrule
			$(m, n)$ & dimension of each patch \tabularnewline \midrule
			$(s, t)$ & dimension of the patch grid \tabularnewline \midrule
			$l_{ij}$ & the likelihood function of the grid cell on $i$th row and $j$th column \tabularnewline
			\bottomrule
	\end{tabular}}
	\label{table::symbols}
\end{table}

First, an input image of size $(h, w)$ will be divided into patches of size $(m, n)$. If the shapes are not divisible, the image will be cropped to the nearest multipliers of each dimension. Therefore, an image of size $(h, w)$ will be divided into a patch matrix of size $(\lfloor h/m \rfloor ,\lfloor w/n \rfloor)$.

Then, the patch matrix will be split into a rectangular patch grid of size $(s, t)$, where each cell contains a certain number of patches. The number of patches in most cells is
\begin{align*}
\left\lfloor \frac{\lfloor h/m \rfloor}{s} \right\rfloor  \times \left\lfloor \frac{\lfloor w/n \rfloor}{t} \right\rfloor,
\end{align*}
except for those cells on the edges, which may have fewer patches.

For each cell in the grid, we fit an outlier detector that is capable of telling the likelihood of a new sample being an outlier. Given a patch $\bm{p}$, it can be reinterpreted by a vector $\bm{v}$, which is given by
\begin{gather*}
\bm{v}=(l_{11}(\bm{p}),~l_{12}(\bm{p}),~l_{13}(\bm{p}),~\ldots,~l_{1t}(\bm{p}),\phantom{).}\\
\phantom{v=(}l_{21}(\bm{p}),~l_{22}(\bm{p}),~l_{23}(\bm{p}),~\ldots,~l_{2t}(\bm{p}),\phantom{).}\\
\phantom{v=(}l_{31}(\bm{p}),~l_{32}(\bm{p}),~l_{33}(\bm{p}),~\ldots,~l_{3t}(\bm{p}),\phantom{).}\\
\cdots\\
\phantom{v=(}l_{s1}(\bm{p}),~l_{s2}(\bm{p}),~l_{s2}(\bm{p}),~\ldots,~l_{st}(\bm{p})).
\end{gather*}

\begin{figure}[htpb]
	\centering
	\includegraphics[width=0.95\linewidth]{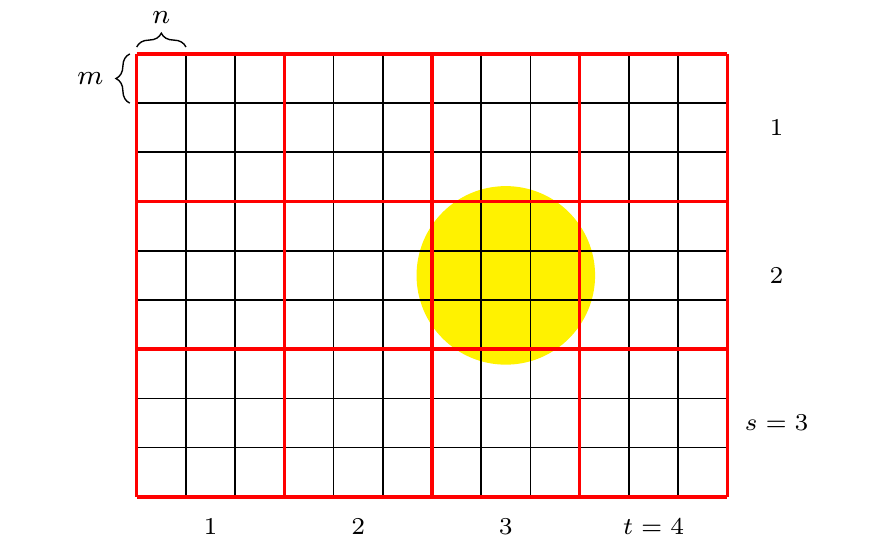}
	\caption{\csentence{Patch reinterpretation illustration.} The parameters are described in Table \ref{table::symbols}.}
	\label{fig::feature_ext_illu}
\end{figure}

An illustration of this reinterpretation method is shown in Figure \ref{fig::feature_ext_illu}, where black blocks represent patches, red blocks represent grid cells and the yellow region represents the tampered region. In this case, $(s, t) = (3, 4)$. Because the tampered region has a different residual pattern, and its contaminated patches concentrate in one of the cells, the outlier detector of that cell will learn a distinct decision boundary compared to other ones. As a result, an authentic patch $\bm{p}_a$ will have lower outlier likelihood in all components except for $l_{23}(\bm{p}_a)$; a tampered patch $\bm{p}_t$ will have higher outlier likelihood in all components except for $l_{23}(\bm{p}_t)$. This difference in structure allows us to distinguish between authentic and tampered patches. In practice, we use the histogram of $\bm{v}$ (denoted by $\bm{v}_h$), which not only encodes the structure in summary-statistics space, but also becomes position invariant. 

\subsubsection{Feature Design}

Besides $\bm{v}_h$, we include some other information in order to concentrate more global information within the feature. The final feature of a patch contains the following components:

\begin{enumerate}
	\item $\bm{v}_h$: the histogramed patch reinterpretation. After generating all histogramed reinterpretations of an image, we normalize them to $[0, 1]$.
	\item Proximity information: how much the patch differs from its neighborhood. We choose the Euclidean distance between the histogramed reinterpretation of the patch and those of its surrounding neighbors'.
	\item Global information: how much the patch differs from the entire image. After computing the histogramed reinterpretations for all patches within an image, we apply $k$-means clustering on them, which generates a set of weights and cluster centroids. The additional global information of a patch is given by the Euclidean distance between the reinterpretation and the cluster centroids, as well as the corresponding weights of the centroids.
\end{enumerate}

\section*{Experiments}\label{sec::experiments}
Due to the lack of science-specific image manipulation detection databases, we synthesize our own database for the experiments.

\subsection*{Datasets}

Our novel scientific image manipulation datasets mainly consist of the following three types of manipulations:

\begin{enumerate}
	\item Removal: covering an image region with a single color or with noise. We manually select a rectangular region to be removed from the image. Then we select another rectangular region to sample the color or noise to fill the removal region, where we can compute the mean $\mu$ and standard deviation $\sigma$ of the pixels. We generate four images for each pair of selection
	according to the configuration given in Table \ref{table::removal_config}.
	
	\begin{table}[htpb]
		\caption{Image generation configuration of removals. There mean of the removal region is equal to that of the sample region's, but we vary the standard deviation from zero (pure color) to two standard deviations to create different visual effects.}
		\centering{\scriptsize
			\begin{tabular}{cc}
				\toprule
				\makecell{Removal Region\\ Mean} & \makecell{Removal Region\\ Standard Deviation}\\ \midrule
				$\mu$ & 0\\
				$\mu$ & $0.5\sigma$\\
				$\mu$ & $\sigma$\\
				$\mu$ & $2\sigma$\\
				\bottomrule
		\end{tabular}}
		\label{table::removal_config}
	\end{table}
	
	\item Splicing: copying content from another image.  We randomly choose a small region from the foreground image and paste it at an arbitrary location on the background image. To create noise inconsistency, the region will either be recompressed with JPEG or processed with sharpening filters.
	
	\item Retouching: modifying the content of the image. We will randomly choose a small region within an image and apply Gaussian blurring to it.
	
\end{enumerate}
These manipulations are selected because we believe that they are more prevalent in problematic scientific papers.

We build two datasets that contain western blot images and microscopy images, respectively. We choose images around these two topics because of their frequency in the literature huge, and they are more susceptible to manipulation. We also create a natural image dataset to compensate for the lack of microscopy images for training. It is only used in the training phase. The details of datasets are shown in Table \ref{table::sci_img_dataset}. The meanings of tampering type abbreviations are shown in Table \ref{table::type_abbrev}.

\begin{table}[htpb]
	\caption{Tampering type abbreviations}
	\centering
	{\scriptsize
		\begin{tabular}{cc}
			\toprule
			Abbreviation & Meaning \\ \midrule
			R & removal images\\
			J & splicing images recompressed by JPEG\\
			F & splicing images processed with sharpening filters\\
			B & Gaussian blurred images\\
			G & genuine images \\ \bottomrule
		\end{tabular}
	}
	
	\label{table::type_abbrev}
\end{table}

\begin{table}[htpb]
	\caption{The specification of the proposed scientific image forensics datasets.}
	\centering{\scriptsize
		\begin{tabular}{C{0.15\linewidth}m{0.25\linewidth}C{0.2\linewidth}c}
			\toprule
			Collection & \makecell[c]{Image\\ Source} & Contents\footnotemark & \makecell[c]{Average\\ Resolution}\\ \midrule
			
			western blot & western blot images from the Internet & R(436), G(51) & $137,244$\\ \midrule
			
			microscopy & microscopy images from the Internet &  R(180), J(20), F(19), B(20), G(21) & $591,906$\\ \midrule
			
			natural image & natural images from the ``pristine'' collection of IEEE dataset \cite{ieeedataset} & J(40), F(40), B(40), G(40) & $775,328$ \\ \bottomrule
		\end{tabular}
	}
	\label{table::sci_img_dataset}
\end{table}
\footnotetext{format: type(number of images)}

\subsection{Test Configurations}
The sizes of images in the western blot collection are significantly smaller. Therefore, we need to train a special model for them. For the microscopy model, we added natural images into the training set to compensate for the lack of data. The patches from residual images are transformed into frequency domain by Discrete Cosine Transform (DCT) because it yields slightly better performance. Within each model, the parameters of each feature extractor are the same. Detailed configurations of the two models that we trained are shown in Table \ref{table::exp_config}.

We use a one-class SVM outlier detector \cite{scholkopf2001estimating}, provided by scikit-learn \cite{scikit-learn}, which is based on LIBSVM\cite{changlibsvm}. The kernel we use is radial basis function, whose kernel coefficient ($\gamma$) is given by the scale, which is
\begin{align*}
\frac{1}{\mbox{number of features} \times \mbox{variance of all inputs}}.
\end{align*}
The tolerance of optimization is set to $0.01$; and $\nu$ (the upper bound on the fraction of training errors and the lower bound on the fraction of support vectors) is set to 0.1.

Note that the choice of parameters can significantly influence the speed of feature extraction. One of the most expensive operations is fitting SVM, which has a computational complexity of $O(N^3)$, where $N$ is the number of patches in each grid cell. Therefore, it is important to choose an appropriate $(m,n)$ and $(s,t)$ pair. With our Python implementation and the configuration given in Table \ref{table::exp_config}, the extraction speed for western blots is approximately 212.36 sec/megapixel (12.13 sec/image), while the extraction speed for microscopy images is approximately 86.15 sec/megapixel (49.32 sec/image). We tried to use ThunderSVM\cite{wenthundersvm18}, which is a GPU-accelerated SVM implementation. Although it has a much higher speed, its precision is not ideal compared to LIBSVM. Therefore, our experiments are conducted with LIBSVM only.

The number of centroids of $k$-means clustering is set to $k=6$, and the clustering algorithm is run 150 times with different initializations in order to get a best result. We select this particular value of $k$ because when we apply $k$-means clustering to $\bm{v}_h$, the tampered region would usually blend with other clusters unless there are more than 6 centroids. Therefore, we consider it reasonable to represent the major content of an image by its first 6 cluster centroids.

\begin{table}[htpb]
	\caption{Test configuration parameters}
	\centering
	{\scriptsize
		\begin{tabular}{ccccc}
			\toprule
			& \makecell[c]{Patch\\ Dimension} & \makecell[c]{Patch Grid\\ Dimension} & \makecell[c]{\# Training\\ Images} & \makecell[c]{\# Testing\\ Images}\\ \cmidrule{2-5}
			Western Blot & $(6, 6)$ & $(5, 5)$ & 352 & 135 \\ \cmidrule{2-5}
			\makecell[c]{Microscopy} & $(10, 10)$ & $(7, 7)$ & 251 & 106 \\ \bottomrule
		\end{tabular}
		\label{table::exp_config}
	}
\end{table}

Because the dimensionality of the extracted feature is not very high, the outputs of each feature extractor are simply concatenated into a single feature vector and then fed to the classifier. The classifier we use is a simple Multilayer Perceptron neural network. For the western blot model, we use a four-layer network with 200 units per layer; for the microscopy model, we use a similar network with 300 units per layer. Softmax regression is applied to the last layer to get the classification results.

\section{Results}

The performance evaluation metric that we use are patch-level accuracy, AUC scores, and F1 scores. We compare the performance of our model with two baseline models, which are widely compared against in related papers:

\begin{enumerate}
	\item CFA \cite{dirik2009image}: a method that uses nearby pixels to evaluate the Camera Filter Array patterns and then produces the tampering probability based on the prediction error.
	\item NOI \cite{mahdian2009using}: a method that finds noise inconsistencies by using high pass wavelet coefficients to model local noise.
\end{enumerate}

For our method, the threshold for F1 score is 0.5. For the baseline methods, their output map is normalized to $[0, 1]$, and the F1 score is acquired by setting the threshold to 0.5. 

\begin{figure*}[p]
	\includegraphics[width=\linewidth]{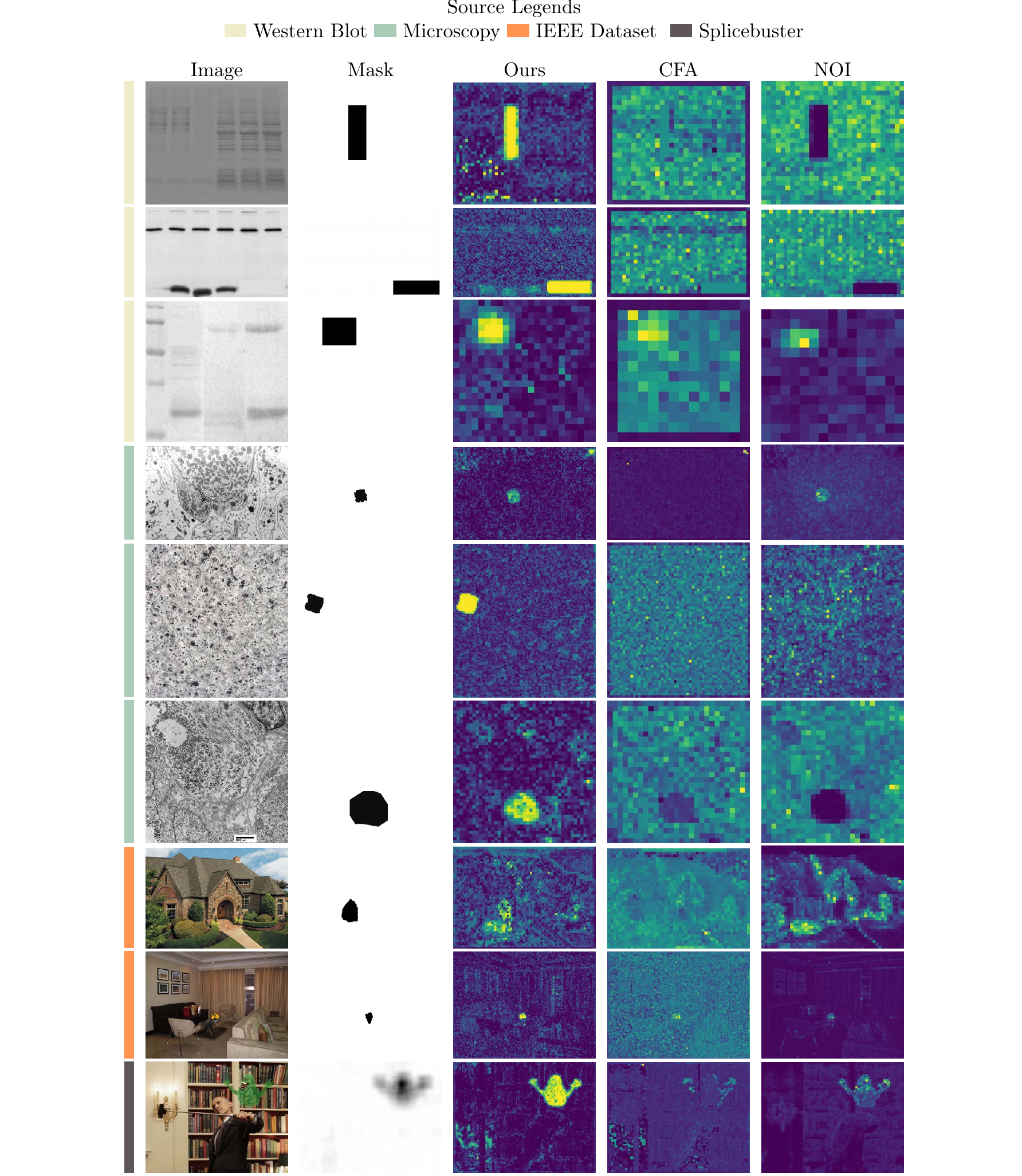}
	\caption{\csentence{Visual comparison of the results.}}
	\label{fig::visual_result}
\end{figure*}

Table \ref{table::gen_acc} shows the accuracies of the three methods on genuine images, where AUC and F1 scores does not apply. Table \ref{table::aucs} and \ref{table::f1s} shows the AUC scores and F1 scores of our methods compared to the baseline. The meanings of the abbreviations can be seen in Table \ref{table::type_abbrev}. The ``overall'' scores are computed across the entire dataset, including genuine images. A visual comparison of the results of each method is shown in Figure \ref{fig::visual_result}. 

It can be seen that CFA cannot handle western blot images very well, as it has low accuracy on genuine images. Its performance on J, F and B tampering types are also mediocre. NOI has better behavior at locating noisy regions in the image, but it fails drastically when encountering manipulations that contain less noise. It constantly treats R[0] and B manipulations as negatives, which yields a false negative region that is not always separable. Its performance on J images is not very satisfactory as well. Generally speaking, the performance of our method is more consistent across different types of manipulations, which makes it more reliable in practice.

\begin{table*}[htpb]
	\caption{The AUC scores on datasets}
	\centering
	{\scriptsize
		\begin{tabular}{clccccclccc}
			\toprule
			& \makecell[c]{Tampering\\ Type} & Ours          & CFA   & NOI   &       &  &    \makecell[c]{Tampering\\ Type}     & Ours          & CFA   & NOI    \\ \cmidrule{2-5} \cmidrule{8-11}
			\multirow{8}{*}{\rotatebox{90}{Western Blot}} & R$[0]$\footnotemark & \textbf{0.939} & 0.606 & 0.026 &  & \multirow{8}{*}{\rotatebox{90}{Microscopy}} & R$[0]$ & \textbf{0.924} & 0.780 & 0.027 \\
			& R$[0.5\sigma]$ & 0.861 & 0.866 & \textbf{0.879} &  &                             & R$[0.5\sigma]$ & \textbf{0.903} & 0.925 & 0.887 \\
			& R$[\sigma]$ & 0.923 & 0.877 & \textbf{0.968} &  &                             & R$[\sigma]$ & \textbf{0.968} & 0.940 & 0.959 \\
			& R$[2\sigma]$  & 0.990 & 0.885 & \textbf{0.992} &  &                             & R$[2\sigma]$  & 0.966 & 0.937 & \textbf{0.978} \\
			&               &       &       &       &  &                             & J          & \textbf{0.994} & 0.639 & 0.618 \\
			&               &       &       &       &  &                             & F        & 0.868 & 0.629 & \textbf{0.913} \\
			&               &       &       &       &  &                             & B    & \textbf{0.805} & 0.334 & 0.104 \\ \cmidrule{2-5} \cmidrule{8-11}
			& overall       & \textbf{0.927} &   0.813    &  0.696    &  &                             & overall       & \textbf{0.925} &  0.864     &  0.695    \\ \bottomrule
	\end{tabular}}
	
	\label{table::aucs}
\end{table*}
\footnotetext{format: R[noise standard deviation]}

\begin{table*}[htpb]
	\caption{The F1 scores on datasets}
	\centering
	{\scriptsize
		\begin{tabular}{clccccclccc}
			\toprule
			& \makecell[c]{Tampering\\ Type}  & Ours  & CFA   & NOI   &  &                             & \makecell[c]{Tampering\\ Type} & Ours  & CFA   & NOI   \\ \cmidrule{2-5} \cmidrule{8-11} 
			\multirow{8}{*}{\rotatebox{90}{Western Blot}} & R$[0]$ & \textbf{0.834} & 0.039 & 0.003 &  & \multirow{8}{*}{\rotatebox{90}{Microscopy}} & R$[0]$ & \textbf{0.834} & 0.039 & 0.000 \\
			& R$[0.5\sigma]$ & \textbf{0.744} & 0.399 & 0.543 &  &                             & R$[0.5\sigma]$ & \textbf{0.745} & 0.398 & 0.560 \\
			& R$[\sigma]$ & \textbf{0.867} & 0.553 & 0.712 &  &                             & R$[\sigma]$ & \textbf{0.867} & 0.414 & 0.773 \\
			& R$[2\sigma]$  & 0.762 & 0.522 & \textbf{0.880} &  &                             & R$[2\sigma]$  & 0.762 & 0.378 & \textbf{0.896} \\
			&        &  &       &       &  &                             & J          & \textbf{0.966} & 0.038 & 0.045 \\
			&               &       &       &       &  &                             & F        & \textbf{0.623} & 0.139 & 0.360 \\
			&               &       &       &       &  &                             & R    & \textbf{0.476} & 0.016 & 0.001 \\ \cmidrule{2-5} \cmidrule{8-11}
			& overall           &   \textbf{0.770}    &  0.300   &  0.424     &  &                             & overall       & \textbf{0.738} &  0.329     & 0.455    \\ \bottomrule
	\end{tabular}}
	
	\label{table::f1s}
\end{table*}

\begin{table}[htpb]
	\caption{The accuracy scores on genuine images}
	\centering
	{\scriptsize
		\begin{tabular}{ccccccc}
			\toprule
			\multicolumn{3}{c}{Western Blot} &  & \multicolumn{3}{c}{Microscopy} \\ \cmidrule{1-3}\cmidrule{5-7}
			Ours  & CFA   & NOI   &  & Ours  & CFA   & NOI   \\
			\textbf{0.988} & 0.513 & 0.838 &  & \textbf{0.988} & 0.774 & 0.920 \\ \bottomrule
	\end{tabular}}
	
	\label{table::gen_acc}
\end{table}

\section{Conclusion And Discussion}\label{sec::conclusion}

We have proposed a novel image tampering detection method for scientific images, which is based on uncovering noise inconsistencies. We use residual images to exploit the noise pattern of the image, and we develop a new feature extraction technique to lower the dimensionality of the problem so that it can be handled by a light-weight classifier. The method is tested on a new scientific image dataset of western blots and microscopy imagery. Compare to two base line methods popular in the literature, results suggest that our method is capable of detecting various types of image manipulations better and more consistently. Thus, our solution promises to solve an important part of image tampering in science effectively.

There are also some weaknesses in our study. First, our proposed method is tested on a custom database, which only contains a small amount of samples. We only include several types of manipulations in our datasets, which is rather monotonous compared to the space of all possible image tampering techniques. Nonetheless, the choice of these specific image sources and manipulation types is inspired by existing problematic papers. If our method is capable of detecting these manipulations to some extent, we believe that it can make valuable discoveries once put into practice.

Second, we think that noise-inconsistency-based methods do possess certain limitations. For example, not all manipulation will necessarily trigger noise inconsistency; it is also easier for one to hide the noise inconsistency, had he/she known the underlying mechanism of the automatic detector. This kind of adversarial attack, however, is significantly challenging and unlikely to be done by the average scientist. In the future, we want to develop more advanced methods that take both image content and noise pattern into account.

However, our proposed method is one of the first methods that tackles scientific image manipulation directly. Put together in screening pipelines for scientific publications (similar to \cite{acuna2018bioscience}), our method would significantly expand the range of manipulations that could be captured at scale. It also makes predictions based on many types of residuals, which possesses improved robustness. The method a set of easily adjustable parameters, which allows it to be adapted for different fields with less effort and a smaller amount of training data.

We would like to continue extending the database with more images from various disciplines to make it standard and comprehensive, and report test results on the updated version. It is our hope that the datasets that we propose can also be useful for the nascent Computational Research Integrity research area. But we are also facing a major difficulty: there are no openly available datasets on images that actually come from \emph{science} (although see the efforts in \cite{beck2016shaping}). The images that we currently have are collected from the Internet, and form a small but significant portion of images with manipulation issues. Unfortunately, access to problematic scientific images are tend to be removed from the public soon after retraction. So far, neither publishers nor authors are yet willing to share those images for understandable reasons. Hopefully, once scientific image tampering detection methods prove their efficacy, publishers and funders can start to share and create datasets with proper safeguards to check for potential problems during peer review -- similar to how they do it with full-text through the Crossref organization\footnote{\url{https://crossref.org}}.

\section*{Acknowledgment}
Daniel E. Acuna and Ziyue Xiang are funded by the Office of Research Integrity grants \#ORIIR180041 and \#ORIIR19001.


\newcommand{\BMCxmlcomment}[1]{}

\BMCxmlcomment{
	
	<refgrp>
	
	<bibl id="B1">
	<title><p>Photo forensics</p></title>
	<aug>
	<au><snm>Farid</snm><fnm>H</fnm></au>
	</aug>
	<publisher>MIT Press</publisher>
	<pubdate>2016</pubdate>
	</bibl>
	
	<bibl id="B2">
	<title><p>{Image Forgery Detection A survey}</p></title>
	<aug>
	<au><snm>Farid</snm><fnm>H</fnm></au>
	</aug>
	<source>{IEEE SIGNAL PROCESSING MAGAZINE}</source>
	<pubdate>{2009}</pubdate>
	<volume>{26}</volume>
	<issue>{2}</issue>
	<fpage>{16</fpage>
		<lpage>25}</lpage>
	</bibl>
	
	<bibl id="B3">
	<title><p>Science journals crack down on image manipulation</p></title>
	<aug>
	<au><snm>Gilbert</snm><fnm>N</fnm></au>
	</aug>
	<publisher>Nature Publishing Group</publisher>
	<pubdate>2009</pubdate>
	</bibl>
	
	<bibl id="B4">
	<title><p>Shaping Images: Scholarly Perspectives on Image
	Manipulation</p></title>
	<aug>
	<au><snm>Beck</snm><fnm>TS</fnm></au>
	</aug>
	<publisher>Walter de Gruyter GmbH \& Co KG</publisher>
	<pubdate>2016</pubdate>
	</bibl>
	
	<bibl id="B5">
	<title><p>The prevalence of inappropriate image duplication in biomedical
	research publications</p></title>
	<aug>
	<au><snm>Bik</snm><fnm>EM</fnm></au>
	<au><snm>Casadevall</snm><fnm>A</fnm></au>
	<au><snm>Fang</snm><fnm>FC</fnm></au>
	</aug>
	<source>MBio</source>
	<publisher>Am Soc Microbiol</publisher>
	<pubdate>2016</pubdate>
	<volume>7</volume>
	<issue>3</issue>
	<fpage>e00809</fpage>
	<lpage>-16</lpage>
	</bibl>
	
	<bibl id="B6">
	<title><p>Bioscience-scale automated detection of figure element
	reuse</p></title>
	<aug>
	<au><snm>Acuna</snm><fnm>DE</fnm></au>
	<au><snm>Brookes</snm><fnm>PS</fnm></au>
	<au><snm>Kording</snm><fnm>KP</fnm></au>
	</aug>
	<source>bioRxiv</source>
	<publisher>Cold Spring Harbor Laboratory</publisher>
	<pubdate>2018</pubdate>
	<fpage>269415</fpage>
	</bibl>
	
	<bibl id="B7">
	<title><p>Detection of copy-move forgery in digital images using SIFT
	algorithm</p></title>
	<aug>
	<au><snm>Huang</snm><fnm>H</fnm></au>
	<au><snm>Guo</snm><fnm>W</fnm></au>
	<au><snm>Zhang</snm><fnm>Y</fnm></au>
	</aug>
	<source>2008 IEEE Pacific-Asia Workshop on Computational Intelligence and
	Industrial Application</source>
	<pubdate>2008</pubdate>
	<volume>2</volume>
	<fpage>272</fpage>
	<lpage>-276</lpage>
	</bibl>
	
	<bibl id="B8">
	<title><p>Exposing digital forgeries by detecting traces of
	resampling</p></title>
	<aug>
	<au><snm>Popescu</snm><fnm>AC</fnm></au>
	<au><snm>Farid</snm><fnm>H</fnm></au>
	</aug>
	<source>IEEE Transactions on signal processing</source>
	<publisher>IEEE</publisher>
	<pubdate>2005</pubdate>
	<volume>53</volume>
	<issue>2</issue>
	<fpage>758</fpage>
	<lpage>-767</lpage>
	</bibl>
	
	<bibl id="B9">
	<title><p>Fast and reliable resampling detection by spectral analysis of
	fixed linear predictor residue</p></title>
	<aug>
	<au><snm>Kirchner</snm><fnm>M</fnm></au>
	</aug>
	<source>Proceedings of the 10th ACM workshop on Multimedia and
	security</source>
	<pubdate>2008</pubdate>
	<fpage>11</fpage>
	<lpage>-20</lpage>
	</bibl>
	
	<bibl id="B10">
	<title><p>On the role of differentiation for resampling detection</p></title>
	<aug>
	<au><snm>Dalgaard</snm><fnm>N</fnm></au>
	<au><snm>Mosquera</snm><fnm>C</fnm></au>
	<au><snm>P{\'e}rez Gonz{\'a}lez</snm><fnm>F</fnm></au>
	</aug>
	<source>2010 IEEE International Conference on Image Processing</source>
	<pubdate>2010</pubdate>
	<fpage>1753</fpage>
	<lpage>-1756</lpage>
	</bibl>
	
	<bibl id="B11">
	<title><p>Normalized energy density-based forensic detection of resampled
	images</p></title>
	<aug>
	<au><snm>Feng</snm><fnm>X</fnm></au>
	<au><snm>Cox</snm><fnm>IJ</fnm></au>
	<au><snm>Doerr</snm><fnm>G</fnm></au>
	</aug>
	<source>IEEE Transactions on Multimedia</source>
	<publisher>Ieee</publisher>
	<pubdate>2012</pubdate>
	<volume>14</volume>
	<issue>3</issue>
	<fpage>536</fpage>
	<lpage>-545</lpage>
	</bibl>
	
	<bibl id="B12">
	<title><p>Blind authentication using periodic properties of
	interpolation</p></title>
	<aug>
	<au><snm>Mahdian</snm><fnm>B</fnm></au>
	<au><snm>Saic</snm><fnm>S</fnm></au>
	</aug>
	<source>IEEE Transactions on Information Forensics and Security</source>
	<publisher>IEEE</publisher>
	<pubdate>2008</pubdate>
	<volume>3</volume>
	<issue>3</issue>
	<fpage>529</fpage>
	<lpage>-538</lpage>
	</bibl>
	
	<bibl id="B13">
	<title><p>On detection of median filtering in digital images</p></title>
	<aug>
	<au><snm>Kirchner</snm><fnm>M</fnm></au>
	<au><snm>Fridrich</snm><fnm>J</fnm></au>
	</aug>
	<source>Media forensics and security II</source>
	<pubdate>2010</pubdate>
	<volume>7541</volume>
	<fpage>754110</fpage>
	</bibl>
	
	<bibl id="B14">
	<title><p>Robust median filtering forensics using an autoregressive
	model</p></title>
	<aug>
	<au><snm>Kang</snm><fnm>X</fnm></au>
	<au><snm>Stamm</snm><fnm>MC</fnm></au>
	<au><snm>Peng</snm><fnm>A</fnm></au>
	<au><snm>Liu</snm><fnm>KR</fnm></au>
	</aug>
	<source>IEEE Transactions on Information Forensics and Security</source>
	<publisher>IEEE</publisher>
	<pubdate>2013</pubdate>
	<volume>8</volume>
	<issue>9</issue>
	<fpage>1456</fpage>
	<lpage>-1468</lpage>
	</bibl>
	
	<bibl id="B15">
	<title><p>Forensic detection of median filtering in digital
	images</p></title>
	<aug>
	<au><snm>Cao</snm><fnm>G</fnm></au>
	<au><snm>Zhao</snm><fnm>Y</fnm></au>
	<au><snm>Ni</snm><fnm>R</fnm></au>
	<au><snm>Yu</snm><fnm>L</fnm></au>
	<au><snm>Tian</snm><fnm>H</fnm></au>
	</aug>
	<source>2010 IEEE International Conference on Multimedia and Expo</source>
	<pubdate>2010</pubdate>
	<fpage>89</fpage>
	<lpage>-94</lpage>
	</bibl>
	
	<bibl id="B16">
	<title><p>Median filtering detection using edge based prediction
	matrix</p></title>
	<aug>
	<au><snm>Chen</snm><fnm>C</fnm></au>
	<au><snm>Ni</snm><fnm>J</fnm></au>
	</aug>
	<source>International Workshop on Digital Watermarking</source>
	<pubdate>2011</pubdate>
	<fpage>361</fpage>
	<lpage>-375</lpage>
	</bibl>
	
	<bibl id="B17">
	<title><p>Forensic detection of image manipulation using statistical
	intrinsic fingerprints</p></title>
	<aug>
	<au><snm>Stamm</snm><fnm>MC</fnm></au>
	<au><snm>Liu</snm><fnm>KR</fnm></au>
	</aug>
	<source>IEEE Transactions on Information Forensics and Security</source>
	<publisher>IEEE</publisher>
	<pubdate>2010</pubdate>
	<volume>5</volume>
	<issue>3</issue>
	<fpage>492</fpage>
	<lpage>-506</lpage>
	</bibl>
	
	<bibl id="B18">
	<title><p>Detect piecewise linear contrast enhancement and estimate
	parameters using spectral analysis of image histogram</p></title>
	<aug>
	<au><snm>{Yao}</snm><fnm>H.</fnm></au>
	<au><snm>{Wang}</snm><fnm>S.</fnm></au>
	<au><snm>{Zhang}</snm><fnm>X.</fnm></au>
	</aug>
	<source>IET International Communication Conference on Wireless Mobile and
	Computing (CCWMC 2009)</source>
	<pubdate>2009</pubdate>
	<fpage>94</fpage>
	<lpage>97</lpage>
	</bibl>
	
	<bibl id="B19">
	<title><p>Blind forensics of contrast enhancement in digital
	images</p></title>
	<aug>
	<au><snm>Stamm</snm><fnm>M</fnm></au>
	<au><snm>Liu</snm><fnm>KR</fnm></au>
	</aug>
	<source>2008 15th IEEE International Conference on Image Processing</source>
	<pubdate>2008</pubdate>
	<fpage>3112</fpage>
	<lpage>-3115</lpage>
	</bibl>
	
	<bibl id="B20">
	<title><p>Forensic estimation and reconstruction of a contrast enhancement
	mapping</p></title>
	<aug>
	<au><snm>Stamm</snm><fnm>MC</fnm></au>
	<au><snm>Liu</snm><fnm>KR</fnm></au>
	</aug>
	<source>2010 IEEE International Conference on Acoustics, Speech and Signal
	Processing</source>
	<pubdate>2010</pubdate>
	<fpage>1698</fpage>
	<lpage>-1701</lpage>
	</bibl>
	
	<bibl id="B21">
	<title><p>Image partial blur detection and classification</p></title>
	<aug>
	<au><snm>Liu</snm><fnm>R</fnm></au>
	<au><snm>Li</snm><fnm>Z</fnm></au>
	<au><snm>Jia</snm><fnm>J</fnm></au>
	</aug>
	<source>2008 IEEE conference on computer vision and pattern
	recognition</source>
	<pubdate>2008</pubdate>
	<fpage>1</fpage>
	<lpage>-8</lpage>
	</bibl>
	
	<bibl id="B22">
	<title><p>Detection of non-aligned double {JPEG} compression with estimation
	of primary compression parameters</p></title>
	<aug>
	<au><snm>Bianchi</snm><fnm>T</fnm></au>
	<au><snm>Piva</snm><fnm>A</fnm></au>
	</aug>
	<source>2011 18th IEEE International Conference on Image Processing</source>
	<pubdate>2011</pubdate>
	<fpage>1929</fpage>
	<lpage>-1932</lpage>
	</bibl>
	
	<bibl id="B23">
	<title><p>Image forgery localization via block-grained analysis of {JPEG}
	artifacts</p></title>
	<aug>
	<au><snm>Bianchi</snm><fnm>T</fnm></au>
	<au><snm>Piva</snm><fnm>A</fnm></au>
	</aug>
	<source>IEEE Transactions on Information Forensics and Security</source>
	<publisher>IEEE</publisher>
	<pubdate>2012</pubdate>
	<volume>7</volume>
	<issue>3</issue>
	<fpage>1003</fpage>
	<lpage>-1017</lpage>
	</bibl>
	
	<bibl id="B24">
	<title><p>{JPEG} compression history estimation for color images</p></title>
	<aug>
	<au><snm>Neelamani</snm><fnm>R</fnm></au>
	<au><snm>De Queiroz</snm><fnm>R</fnm></au>
	<au><snm>Fan</snm><fnm>Z</fnm></au>
	<au><snm>Dash</snm><fnm>S</fnm></au>
	<au><snm>Baraniuk</snm><fnm>RG</fnm></au>
	</aug>
	<source>IEEE Transactions on Image Processing</source>
	<publisher>IEEE</publisher>
	<pubdate>2006</pubdate>
	<volume>15</volume>
	<issue>6</issue>
	<fpage>1365</fpage>
	<lpage>-1378</lpage>
	</bibl>
	
	<bibl id="B25">
	<title><p>A convolutive mixing model for shifted double {JPEG} compression
	with application to passive image authentication</p></title>
	<aug>
	<au><snm>Qu</snm><fnm>Z</fnm></au>
	<au><snm>Luo</snm><fnm>W</fnm></au>
	<au><snm>Huang</snm><fnm>J</fnm></au>
	</aug>
	<source>2008 IEEE International Conference on Acoustics, Speech and Signal
	Processing</source>
	<pubdate>2008</pubdate>
	<fpage>1661</fpage>
	<lpage>-1664</lpage>
	</bibl>
	
	<bibl id="B26">
	<title><p>Image tamper detection based on demosaicing artifacts</p></title>
	<aug>
	<au><snm>Dirik</snm><fnm>AE</fnm></au>
	<au><snm>Memon</snm><fnm>N</fnm></au>
	</aug>
	<source>2009 16th IEEE International Conference on Image Processing
	(ICIP)</source>
	<pubdate>2009</pubdate>
	<fpage>1497</fpage>
	<lpage>-1500</lpage>
	</bibl>
	
	<bibl id="B27">
	<title><p>Exploring {DCT} coefficient quantization effects for local
	tampering detection</p></title>
	<aug>
	<au><snm>Wang</snm><fnm>W</fnm></au>
	<au><snm>Dong</snm><fnm>J</fnm></au>
	<au><snm>Tan</snm><fnm>T</fnm></au>
	</aug>
	<source>IEEE Transactions on Information Forensics and Security</source>
	<publisher>IEEE</publisher>
	<pubdate>2014</pubdate>
	<volume>9</volume>
	<issue>10</issue>
	<fpage>1653</fpage>
	<lpage>-1666</lpage>
	</bibl>
	
	<bibl id="B28">
	<title><p>Using noise inconsistencies for blind image forensics</p></title>
	<aug>
	<au><snm>Mahdian</snm><fnm>B</fnm></au>
	<au><snm>Saic</snm><fnm>S</fnm></au>
	</aug>
	<source>Image and Vision Computing</source>
	<publisher>Elsevier</publisher>
	<pubdate>2009</pubdate>
	<volume>27</volume>
	<issue>10</issue>
	<fpage>1497</fpage>
	<lpage>-1503</lpage>
	</bibl>
	
	<bibl id="B29">
	<title><p>Rich models for steganalysis of digital images</p></title>
	<aug>
	<au><snm>Fridrich</snm><fnm>J</fnm></au>
	<au><snm>Kodovsky</snm><fnm>J</fnm></au>
	</aug>
	<source>IEEE Transactions on Information Forensics and Security</source>
	<publisher>IEEE</publisher>
	<pubdate>2012</pubdate>
	<volume>7</volume>
	<issue>3</issue>
	<fpage>868</fpage>
	<lpage>-882</lpage>
	</bibl>
	
	<bibl id="B30">
	<title><p>Steganalysis by subtractive pixel adjacency matrix</p></title>
	<aug>
	<au><snm>Pevny</snm><fnm>T</fnm></au>
	<au><snm>Bas</snm><fnm>P</fnm></au>
	<au><snm>Fridrich</snm><fnm>J</fnm></au>
	</aug>
	<source>IEEE Transactions on information Forensics and Security</source>
	<publisher>IEEE</publisher>
	<pubdate>2010</pubdate>
	<volume>5</volume>
	<issue>2</issue>
	<fpage>215</fpage>
	<lpage>-224</lpage>
	</bibl>
	
	<bibl id="B31">
	<title><p>General-purpose image forensics using patch likelihood under image
	statistical models</p></title>
	<aug>
	<au><snm>Fan</snm><fnm>W</fnm></au>
	<au><snm>Wang</snm><fnm>K</fnm></au>
	<au><snm>Cayre</snm><fnm>F</fnm></au>
	</aug>
	<source>2015 IEEE International Workshop on Information Forensics and
	Security (WIFS)</source>
	<pubdate>2015</pubdate>
	<fpage>1</fpage>
	<lpage>-6</lpage>
	</bibl>
	
	<bibl id="B32">
	<title><p>Splicebuster: A new blind image splicing detector</p></title>
	<aug>
	<au><snm>Cozzolino</snm><fnm>D</fnm></au>
	<au><snm>Poggi</snm><fnm>G</fnm></au>
	<au><snm>Verdoliva</snm><fnm>L</fnm></au>
	</aug>
	<source>"\url{http://www.grip.unina.it/index.php?option=com_content&view=article&id=79&Itemid=489&jsmallfib=1&dir=JSROOT/Splicebuster}""</source>
	<pubdate>2015</pubdate>
	<fpage>1</fpage>
	<lpage>-6</lpage>
	<note>(program and test images are downloaded from URL)</note>
	</bibl>
	
	<bibl id="B33">
	<title><p>Exploiting spatial structure for localizing manipulated image
	regions</p></title>
	<aug>
	<au><snm>Bappy</snm><fnm>JH</fnm></au>
	<au><snm>Roy Chowdhury</snm><fnm>AK</fnm></au>
	<au><snm>Bunk</snm><fnm>J</fnm></au>
	<au><snm>Nataraj</snm><fnm>L</fnm></au>
	<au><snm>Manjunath</snm><fnm>BS</fnm></au>
	</aug>
	<source>Proceedings of the IEEE International Conference on Computer
	Vision</source>
	<pubdate>2017</pubdate>
	<fpage>4970</fpage>
	<lpage>-4979</lpage>
	</bibl>
	
	<bibl id="B34">
	<title><p>A deep learning approach to universal image manipulation detection
	using a new convolutional layer</p></title>
	<aug>
	<au><snm>Bayar</snm><fnm>B</fnm></au>
	<au><snm>Stamm</snm><fnm>MC</fnm></au>
	</aug>
	<source>Proceedings of the 4th ACM Workshop on Information Hiding and
	Multimedia Security</source>
	<pubdate>2016</pubdate>
	<fpage>5</fpage>
	<lpage>-10</lpage>
	</bibl>
	
	<bibl id="B35">
	<title><p>Constrained convolutional neural networks: A new approach towards
	general purpose image manipulation detection</p></title>
	<aug>
	<au><snm>Bayar</snm><fnm>B</fnm></au>
	<au><snm>Stamm</snm><fnm>MC</fnm></au>
	</aug>
	<source>IEEE Transactions on Information Forensics and Security</source>
	<publisher>IEEE</publisher>
	<pubdate>2018</pubdate>
	<volume>13</volume>
	<issue>11</issue>
	<fpage>2691</fpage>
	<lpage>-2706</lpage>
	</bibl>
	
	<bibl id="B36">
	<title><p>Learning Rich Features for Image Manipulation Detection</p></title>
	<aug>
	<au><snm>Zhou</snm><fnm>P</fnm></au>
	<au><snm>Han</snm><fnm>X</fnm></au>
	<au><snm>Morariu</snm><fnm>VI</fnm></au>
	<au><snm>Davis</snm><fnm>LS</fnm></au>
	</aug>
	<source>arXiv preprint arXiv:1805.04953</source>
	<pubdate>2018</pubdate>
	</bibl>
	
	<bibl id="B37">
	<title><p>A universal image forensic strategy based on steganalytic
	model</p></title>
	<aug>
	<au><snm>Qiu</snm><fnm>X</fnm></au>
	<au><snm>Li</snm><fnm>H</fnm></au>
	<au><snm>Luo</snm><fnm>W</fnm></au>
	<au><snm>Huang</snm><fnm>J</fnm></au>
	</aug>
	<source>Proceedings of the 2nd ACM workshop on Information hiding and
	multimedia security</source>
	<pubdate>2014</pubdate>
	<fpage>165</fpage>
	<lpage>-170</lpage>
	</bibl>
	
	<bibl id="B38">
	<title><p>A Picture’s Worth...</p></title>
	<aug>
	<au><snm>Krawetz</snm><fnm>N</fnm></au>
	</aug>
	<source>Hacker Factor Solutions</source>
	<pubdate>2007</pubdate>
	<volume>6</volume>
	</bibl>
	
	<bibl id="B39">
	<title><p>Estimating the support of a high-dimensional
	distribution</p></title>
	<aug>
	<au><snm>Sch{\"o}lkopf</snm><fnm>B</fnm></au>
	<au><snm>Platt</snm><fnm>JC</fnm></au>
	<au><snm>Shawe Taylor</snm><fnm>J</fnm></au>
	<au><snm>Smola</snm><fnm>AJ</fnm></au>
	<au><snm>Williamson</snm><fnm>RC</fnm></au>
	</aug>
	<source>Neural computation</source>
	<publisher>MIT Press</publisher>
	<pubdate>2001</pubdate>
	<volume>13</volume>
	<issue>7</issue>
	<fpage>1443</fpage>
	<lpage>-1471</lpage>
	</bibl>
	
	<bibl id="B40">
	<title><p>Scikit-learn: Machine Learning in {P}ython</p></title>
	<aug>
	<au><snm>Pedregosa</snm><fnm>F.</fnm></au>
	<au><snm>Varoquaux</snm><fnm>G.</fnm></au>
	<au><snm>Gramfort</snm><fnm>A.</fnm></au>
	<au><snm>Michel</snm><fnm>V.</fnm></au>
	<au><snm>Thirion</snm><fnm>B.</fnm></au>
	<au><snm>Grisel</snm><fnm>O.</fnm></au>
	<au><snm>Blondel</snm><fnm>M.</fnm></au>
	<au><snm>Prettenhofer</snm><fnm>P.</fnm></au>
	<au><snm>Weiss</snm><fnm>R.</fnm></au>
	<au><snm>Dubourg</snm><fnm>V.</fnm></au>
	<au><snm>Vanderplas</snm><fnm>J.</fnm></au>
	<au><snm>Passos</snm><fnm>A.</fnm></au>
	<au><snm>Cournapeau</snm><fnm>D.</fnm></au>
	<au><snm>Brucher</snm><fnm>M.</fnm></au>
	<au><snm>Perrot</snm><fnm>M.</fnm></au>
	<au><snm>Duchesnay</snm><fnm>E.</fnm></au>
	</aug>
	<source>Journal of Machine Learning Research</source>
	<pubdate>2011</pubdate>
	<volume>12</volume>
	<fpage>2825</fpage>
	<lpage>-2830</lpage>
	</bibl>
	
	<bibl id="B41">
	<title><p>{LIBSVM}: A library for support vector machines</p></title>
	<aug>
	<au><snm>Chang</snm><fnm>CC</fnm></au>
	<au><snm>Lin</snm><fnm>CJ</fnm></au>
	</aug>
	<source>ACM Transactions on Intelligent Systems and Technology</source>
	<pubdate>2011</pubdate>
	<volume>2</volume>
	<fpage>27:1</fpage>
	<lpage>-27:27</lpage>
	<note>Software available at
	\url{http://www.csie.ntu.edu.tw/~cjlin/libsvm}</note>
	</bibl>
	
	<bibl id="B42">
	<title><p>{ThunderSVM}: A Fast {SVM} Library on {GPUs} and {CPUs}</p></title>
	<aug>
	<au><snm>Wen</snm><fnm>Z</fnm></au>
	<au><snm>Shi</snm><fnm>J</fnm></au>
	<au><snm>Li</snm><fnm>Q</fnm></au>
	<au><snm>He</snm><fnm>B</fnm></au>
	<au><snm>Chen</snm><fnm>J</fnm></au>
	</aug>
	<source>Journal of Machine Learning Research</source>
	<pubdate>2018</pubdate>
	<volume>19</volume>
	<fpage>1</fpage>
	<lpage>-5</lpage>
	</bibl>
	
	<bibl id="B43">
	<title><p>IEEE IFS-TC Image Forensics Challenge Dataset</p></title>
	<aug>
	<au><cnm>{IEEE Information Forensics and Security Technical
		Committee}</cnm></au>
	</aug>
	<source>"\url{http://ifc.recod.ic.unicamp.br/fc.website/index.py?sec=5}""</source>
	<pubdate>2013</pubdate>
	</bibl>
	
	</refgrp>
} 

\end{document}